\documentclass[runningheads]{llncs}

\usepackage{graphicx}
\usepackage{multirow}
\usepackage{amsmath}
\usepackage[linesnumbered,lined,boxed,commentsnumbered]{algorithm2e}
\usepackage{booktabs}
\usepackage{xspace}
\usepackage{siunitx}
\usepackage{hyperref}
\usepackage{tabularx}
\usepackage{svg}

\begin{document}

\title{Unit-Aware Genetic Programming for the Development of Empirical Equations}

\titlerunning{Unit-aware Genetic Programming for Empirical Equations}

\author{Julia Reuter \inst{1}\orcidID{0000-0002-7023-7965} \and
Viktor Martinek \inst{2}\orcidID{0000-0001-6215-4783} \and Roland Herzog \inst{2}\orcidID{0000-0003-2164-6575} \and
Sanaz Mostaghim \inst{1}\orcidID{0000-0002-9917-5227}}
\authorrunning{J. Reuter et al.}

\institute{
{Faculty of Computer Science, Otto-von-Guericke-Universität Magdeburg, Germany}\and {Interdisciplinary Center for Scientific Computing, Universität Heidelberg, Germany}
 	\\
 \email{\{julia.reuter | sanaz.mostaghim\}@ovgu.de \\
 \{viktor.martinek | roland.herzog\}@iwr.uni-heidelberg.de}
}
\maketitle              

\begin{abstract}
When developing empirical equations, domain experts require these to be accurate and adhere to physical laws.  
Often, constants with unknown units need to be discovered alongside the equations. 
Traditional unit-aware genetic programming (GP) approaches cannot be used when unknown constants with undetermined units are included. 
This paper presents a method for dimensional analysis that propagates unknown units as “jokers” and returns the magnitude of unit violations. 
We propose three methods, namely evolutive culling, a repair mechanism, and a multi-objective approach, to integrate the dimensional analysis in the GP algorithm. 
Experiments on datasets with ground truth demonstrate comparable performance of evolutive culling and the multi-objective approach to a baseline without dimensional analysis. 
Extensive analysis of the results on datasets without ground truth reveals that the unit-aware algorithms make only low sacrifices in accuracy, while producing unit-adherent solutions. 
Overall, we presented a promising novel approach for developing unit-adherent empirical equations.

\end{abstract}
\section{Introduction}

Lately, the need to analyze and understand the behavior of machine learning (ML) models has increased to gain a more profound understanding of the underlying system, and to avoid unexpected behaviors.
To this end, combining ML techniques with physics principles is a promising approach.
For data-driven methods such as deep learning, various methods have been proposed to enforce desired behaviors in the models \cite{karniadakisPhysicsinformedMachineLearning2021}.
However, it is impossible to test the entire possible input space of variables to such a system, so that some level of uncertainty in the behavior will always remain.

Symbolic regression (SR) algorithms, on the other hand, produce free-form equations from data.
These allow domain experts to analyze the behavior of the underlying systems.
For many engineering and physics applications, such equations are only trustworthy and useful if they reflect certain physical properties.

Genetic programming (GP) from the family of evolutionary algorithms (EA) is an established method for SR.
Unit-aware GP is an approach to encourage the compliance with physical laws.
The goal is to produce equations that account for input variable units, adhere to physical laws during computation, and yield the same physical unit in the output as the target variable.
Our previous work shows that the optimization process can benefit from unit information of input and target variables \cite{reuterMultiObjectiveIslandModel2023,zilleASSESSMENTMULTIOBJECTIVECOEVOLUTIONARY2021}.
These papers address benchmark equations without any constants.
Other publications treat constants as variables with a constant value and known units.
In both scenarios, the known units can guide the algorithm towards the correct solution.

In practice, the identification of new empirical equations is more complex: the amount, value, and position within the equation as well as the units of the constants are unknown.
Every constant can take on arbitrary units, which makes previous unit-aware GP approaches impossible to use.
However, both, unknown constants and symbolic models that adhere to physical unit constraints, are an important requirement of domain experts from various scientific fields.

In this paper, we apply a method for dimensional analysis that includes constants with undetermined units similar to \texttt{SymbolicRegression.jl}.
The unit of a constant is treated as a “joker”, which can take on any unit.
We propose different techniques to handle unit violations, from the area of constraint handling and multi-objective optimization.
We study the effect of these techniques using datasets of equations that have been discovered empirically in the past.
We also apply different noise levels to the benchmark datasets to examine how sensitive our approaches are to noisy data.
We know from related studies that the importance of prior knowledge increases as the noise in the data increases \cite{haiderShapeConstrainedSymbolicRegression2022a,keijzerDimensionallyAwareGenetic1999a}.
This paper intends to study whether this effect is also observable when constants with unknown units are used.
We furthermore test the proposed methods on datasets without ground truth from fluid mechanics and thermodynamics.
Our experiments indicate that unit-adherent equations can be as accurate as others.
Our research contributes to investigations on integrating domain knowledge into GP algorithms to generate useful solutions for domain experts.

\section{Background and Related Work}
Increasing attention is given to integrating prior knowledge into data-driven modeling, with recent papers specifically addressing this aspect.
SR methods generally have a large search space of possible equations, especially as the complexity of the searched functions increases.
This often leads to problems such as convergence to local optima, overfitting, or loss of interpretability.
The main motivation to exploit prior knowledge is to reduce the search space and guide the search towards useful models.

\subsection{Integration of Physics and Prior Knowledge in SR Algorithms}

A prominent technique for the identification of symbolic models for dynamical systems using prior knowledge is \textit{sparse identification of nonlinear dynamics (SINDY)} \cite{bruntonDiscoveringGoverningEquations2016}.
It uses sparse regression on a function basis of selected functional terms which appear frequently in governing equations of dynamical systems.
Various publications demonstrate the success of the method, even for long-standing problems in science \cite{bruntonDiscoveringGoverningEquations2016,desilvaDiscoveryPhysicsData2020,kaptanogluPySINDyComprehensivePython2022}.
The applicability is, however, limited to identifying models that exclusively consist of the functional terms provided.  

AI Feynman is another physics-inspired method for symbolic regression \cite{udrescuAIFeynmanPhysicsinspired2020}.
The goal is to identify functions of practical interest, which often share certain characteristics such as symmetries, separability, as well as consistency in terms of physical units.
A dimensional analysis component takes the units of the variables into account and matches combinations of these variables with a given target unit.
This approach requires all units to be known in advance. 
This is not the case when searching for new empirical equations with unknown constants.
However, the dimensional analysis component can be considered a counter-movement to contemporary machine learning methods, which often standardizes features into dimensionless quantities.
AI Feynman shows that, indeed, unit information can be valuable to the algorithm.

In \cite{kerenComputationalFrameworkPhysicsinformed2023}, the applicability of existing SR methods for physical systems is discussed.
The \texttt{SciMED} framework is proposed, a scientist-in-the-loop approach, to include prior knowledge in the search for useful equations.
Their method outperforms AI Feynman as well as GP-GOMEA~\cite{virgolinImprovingModelBasedGenetic2021} in some cases.
Generally, GP-based SR approaches provide the opportunity to include prior knowledge on various levels.
Popular frameworks such as \texttt{PySR} allow for user-defined functions, additional objectives or certain building rules \cite{cranmerInterpretableMachineLearning2023}.
The inclusion of shape constraints in GP algorithms as additional objectives was extensively studied in \cite{haiderShapeconstrainedMultiobjectiveGenetic2023a,haiderShapeConstrainedSymbolicRegression2022a}.
Overall, the benefit of knowledge about target functional shapes increases with the noise level in the training data.

\subsection{Unit-aware Genetic Programming}
\label{sec:unitawareGP}
The consideration of physical units in the search for symbolic models was studied early in the GP area.
Keijzer and Babovic suggested different methods to handle unit violations in GP \cite{keijzerDimensionallyAwareGenetic1999a}.
A multi-objective approach minimizing the dimension error yields the best results, and unit information gains importance as the noise level of the data increases.
The algorithms with dimensional analysis only found the ground truth solutions regularly when the used constants and units were given as input features.
This approach sets the foundations of our work.
However, contrary to \cite{keijzerDimensionallyAwareGenetic1999a}, our approach does not assume that new constants are dimensionless, which makes a big difference in the dimensional analysis.

Some methods from the literature prevent the generation of invalid individuals by defining a building grammar, which was used for unit-aware feature construction for experimental physics \cite{cherrierConsistentFeatureConstruction2019} or construction of multigrid solvers \cite{schmittEvoStencilsGrammarbasedGenetic2021}.
Others allow the building of solutions with unit violations, and define methods to handle them.
For dimensional analysis with undetermined units of constants, we see the latter as more feasible.
Overall, we can identify three predominant ways to deal with unit-related constraint violations in GP: 
first, a multi-objective variant that minimizes the number of unit violations as an additional objective \cite{liDimensionallyAwareMultiObjective2020,zilleASSESSMENTMULTIOBJECTIVECOEVOLUTIONARY2021}. Second, a correction mechanism that manipulates the model to match the input and target unit, for example by multiplication with a constant \cite{keijzerDimensionallyAwareGenetic1999a}.
And third, the addition of a penalty term for unit violations to the primary objective \cite{cranmerInterpretableMachineLearning2023}.
The most drastic case is the “death penalty”, which assigns a large penalty value to guarantee that an individual will not survive to the next generation \cite{bandaruDimensionallyAwareGeneticProgramming2013,meiConstrainedDimensionallyAware2017}.
The brood selection strategy by \cite{keijzerDimensionallyAwareGenetic1999a} has similarities with the death penalty approach: multiple offspring are generated from one individual, and the one with the smallest unit violation will be added to the population.
It is applied already at the reproduction and not at the selection stage of an algorithm.

The \texttt{PySR} backend \texttt{SymbolicRegression.jl} recently released a functionality to consider unknown constants in the dimensional analysis \cite{cranmerInterpretableMachineLearning2023}.
The equation is evaluated, and the units are propagated through the equation accordingly.
Constants act as so-called “wildcards” and can take on arbitrary units.
In case of unit violations, a penalty term is added to the primary objective.
This penalty does not account for the number of unit violations, i.e., solutions with few violations are treated equally to solutions with many violations.
Depending on the penalty value, this can have the effect of a death penalty.

In this paper, we assess different methods to handle unit violations using a dimensional analysis function that accounts for unknown constants.
We use the unit propagation scheme from \texttt{SymbolicRegression.jl} as a starting point for our implementations.
Our approach considers the number of unit violations in the dimensional analysis, rather than returning a boolean value that indicates whether a violation occurs.
Furthermore, we propose and compare different ways to account for unit violations in the evolutionary process.
Combining parameter estimation with the death penalty for constraint handling has the negative effect that a solution, that does not survive to the next generation because of the death penalty, still uses computational resources for the expensive parameter estimation.
Our proposed constraint handling approaches exploit the cheaper dimensional analysis to handle unit violations before fitting. We furthermore assess a multi-objective approach, considering the magnitude of unit violations.

\section{Unit-aware Genetic Programming with Unknown Constants}

Genetic Programming for SR is a well-established population-based approach to develop symbolic models from data \cite{kozaGeneticProgrammingMeans1994}.
Equations are usually represented as trees, which are formed using elements from the feature and function sets.
Starting from an initial random population of trees, crossover and mutation operations are applied iteratively to create new individuals.
For SR tasks, the equations are typically evaluated using the prediction error on the target variable as the primary fitness measure.
A complexity measure is included as a second optimization criterion to avoid bloat and present a set of Pareto-optimal (PO) solutions in terms of error and complexity to the decision maker.

\subsection{Dimensional Analysis with Unknown Constants}

In GP, different types of constants are used in practice.
When only known constants are used, they can be included in the training data and treated like regular features (e.g., the Feynman datasets \cite{udrescuAIFeynmanPhysicsinspired2020}).
When unknown constants are used, contemporary GP-based SR methods use parameter estimation on top of the evolutionary process.
The number and position of constants within an equation is determined during the generation of a tree.
The values of the constants are then fitted to the target variable using a parameter estimation algorithm.
This fitting process is a computationally expensive task.

\renewcommand{\arraystretch}{1.1}
\begin{table}[t!]
    \centering
    \caption{Set of common functions used in GP with their expected input units and the resulting output unit. A joker unit is represented as [$\diamondsuit, \diamondsuit, \diamondsuit$].}
        \begin{tabular}{c|c|c}
         Function &  Units of operands & Unit after execution of function\\ \hline
         \multirow{ 3}{*}{$+$, $-$} & [$a, b, c$], [$a, b, c$] & [$a, b, c$] \\   \cline{2-3}
         & [$a, b, c$], [$\diamondsuit, \diamondsuit, \diamondsuit$] & [$a, b, c$] \\  \cline{2-3}
         & [$\diamondsuit, \diamondsuit, \diamondsuit$], [$\diamondsuit, \diamondsuit, \diamondsuit$] & [$\diamondsuit, \diamondsuit, \diamondsuit$] \\\hline

         \multirow{ 3}{*}{$\cdot$} & [$a, b, c$], [$d, e, f$]& [$a+d, b+e, c+f$] \\\cline{2-3}
         & [$a, b, c$], [$\diamondsuit, \diamondsuit, \diamondsuit$] & [$\diamondsuit, \diamondsuit, \diamondsuit$] \\ \cline{2-3}
         & [$\diamondsuit, \diamondsuit, \diamondsuit$], [$\diamondsuit, \diamondsuit, \diamondsuit$] & [$\diamondsuit, \diamondsuit, \diamondsuit$] \\\hline

         \multirow{ 3}{*}{$\div$} & [$a, b, c$], [$d, e, f$]& [$a-d, b-e, c-f$] \\\cline{2-3}
          & [$a, b, c$], [$\diamondsuit, \diamondsuit, \diamondsuit$] & [$\diamondsuit, \diamondsuit, \diamondsuit$] \\ \cline{2-3}
          & [$\diamondsuit, \diamondsuit, \diamondsuit$], [$\diamondsuit, \diamondsuit, \diamondsuit$] & [$\diamondsuit, \diamondsuit, \diamondsuit$] \\\hline

         \multirow{ 2}{*}{$e^{\circ}$, $\log({\circ})$} & [$0, 0, 0$] & [$0, 0, 0$]\\ \cline{2-3}
          & [$\diamondsuit, \diamondsuit, \diamondsuit$] & [0, 0, 0]\\\hline

         \multirow{ 2}{*}{$\sin(\circ)$, $\cos(\circ)$, $\tan(\circ)$} & [$0, 0, 0$] & [$0, 0, 0$] \\ \cline{2-3}
          & [$\diamondsuit, \diamondsuit, \diamondsuit$] & [$0, 0, 0$] \\ \hline

         \multirow{ 2}{*}{$\sqrt{\circ}$}  & [$a, b, c$] & [$\frac{a}{2}, \frac{b}{2}, \frac{c}{2}$]\\ \cline{2-3}
          & [$\diamondsuit, \diamondsuit, \diamondsuit$] & [$\diamondsuit, \diamondsuit, \diamondsuit$]\\ \hline

         \multirow{ 2}{*}{$\circ^k$ (k $\in$ $\mathbf{N}$)} & [a, b, c] & [a $\cdot$ k, b $\cdot$ k, c $\cdot$ k] \\ \cline{2-3}
         & [$\diamondsuit, \diamondsuit, \diamondsuit$] & [$\diamondsuit, \diamondsuit, \diamondsuit$]  \\ \hline

         \multirow{ 3}{*}{$\circ^{\circ}$ (binary power operator)} & [$0, 0, 0$], [$0, 0, 0$] & [$0, 0, 0$]  \\ \cline{2-3}
          & [$\diamondsuit, \diamondsuit, \diamondsuit$], [$0, 0, 0$] & [$0, 0, 0$] \\  \cline{2-3}
          & [$\diamondsuit, \diamondsuit, \diamondsuit$], [$\diamondsuit, \diamondsuit, \diamondsuit$] & [$0, 0, 0$] \\
    \end{tabular}
    \label{tab:my_label}
\end{table}
We express the units of a variable as a vector of exponents of SI units with the order [\unit{\metre}, \unit{\kilogram}, \unit{\second}, \unit{\ampere}, \unit{\kelvin}, \unit{\mol}, \unit{\candela}].
A quantity in Newton $[\unit{\newton}]$ = [$\frac{\mathrm{kg} \cdot \mathrm{m}}{\mathrm{s}^2}$] can thus be expressed as $[1, 1, -2, 0, 0, 0, 0]$.
In this paper, constants have generally unknown units, which makes traditional approaches to detect unit violations infeasible.
To overcome this issue, we apply the unit propagation scheme, similarly as implemented in \texttt{SymbolicRegression.jl}, and introduce a joker unit [$\diamondsuit, \diamondsuit, \diamondsuit$], representing unknown units.
Dimensionless inputs are expressed as $[0, 0, 0]$.
Table~\ref{tab:my_label} displays how our proposed algorithm handles operands with known and unknown units for a set of functions that are commonly used in SR algorithms.
This set of functions is non-exhaustive and can be extended to custom functions as well.
For the sake of readability, we display only three elements of the unit vector.
The rules, however, apply to all seven elements.

The use of joker units leads to some special cases which need to be addressed: addition and subtraction require equal units of both operands.
If one operand is a joker, the unit of the other operand is returned.
If both operands are jokers, a joker is returned.
For multiplication and division, one or two joker operands produce a joker output.
Functions requiring dimensionless inputs assume that a joker operand is dimensionless, and return a dimensionless quantity accordingly.
Operations with fixed exponents ($\sqrt{\circ}$ and power operations $\circ^2, \circ^3, \dots$) produce a joker output if the function input is a joker.
The binary power operator requires both operands to be dimensionless and returns a dimensionless quantity.
If one or two operands have joker units, they are assumed to be dimensionless to return a dimensionless quantity.


\newcommand\alggray{40}
\newcommand\algcomment[1]{{\itshape// #1}}

\newcommand\nisconstant{$n$ \textnormal{is a constant}$\;$}
\newcommand\nisvariable{$n$ \textnormal{is a variable}$\;$}
\newcommand\nispowerop{$n$ \textnormal{is power operator}$\;$}

\newcommand\nisunary{$n$ \textnormal{is a unary operation}$\;$}
\newcommand\nisbinary{$n$ \textnormal{is a binary operation}$\;$}
\newcommand\ninpm{$n \in \{+,-\}$}
\newcommand\dleft{$d_\textrm{left}$}
\newcommand\dright{$d_\textrm{right}$}
\newcommand\ninpow{n \in \{ \hspace{1mm}\^{} \hspace{1mm} \}}
\newcommand\dldr{d_{\textrm{left}}, d_{\textrm{right}}}
\newcommand\unitsmatch{\textnormal{units match case from Table \ref{tab:my_label}}}
\newcommand\algorithmsize{\small}

\SetStartEndCondition{ }{}{}\SetKwProg{Fn}{function}{}{end}
\SetKwFunction{recDimAnalysis}{recDimAnalysis}
\SetKwFunction{choice}{choice}
\SetKwComment{Comment}{/* }{ */}
\SetKwInOut{Input}{input}
\SetKwInOut{Output}{output}
\SetStartEndCondition{ }{}{}
\SetKwProg{Fn}{function}{}{end}
\SetKwProg{Fn}{function}{\string:}{}
\RestyleAlgo{ruled}
\DontPrintSemicolon

\begin{algorithm*}[t!]
\algorithmsize
\caption{Recursive Dimensional Analysis}\label{alg:dimPen}
\Input{Root node $n$ of the tree, units of variables}
\Output{Tuple (output dimension $d$, number of unit violations $v$)}

\Fn{\recDimAnalysis{$n$}}{
\If{\nisconstant}
{\textbf{return}([$\diamondsuit, \diamondsuit, \diamondsuit$], 0)}

\If{\nisvariable}
{\textbf{return}([$a, b, c$], 0)}

\If{\nisunary}{
\text{$d$, $v$  $\gets$ \recDimAnalysis{$n.child$}}\;

\eIf{\unitsmatch }{
        $d \gets$ unit after execution of operation\;
        }{
        $d \gets$ true output unit of operation\;
        $v \gets v+1$ \;
        }
    \textbf{return} ($d$, $v$)\;}

\If{\nisbinary}{
\text{$d_{\textrm{right}}$, $v_{\textrm{right}}$  $\gets$ \recDimAnalysis{n.right}} \;
\text{$d_{\textrm{left}}$, $v_{\textrm{left}}$  $\gets$ \recDimAnalysis{n.left}} \;

\If{\unitsmatch}{
$d \gets$ unit after execution of operation\;
$v \gets$ $v_{\textrm{right}}$ + $v_{\textrm{left}}$ \;
}

\If{\ninpm}{
$v \gets$ $v_{\textrm{right}}$ + $v_{\textrm{left}}$ + 1 \;
$d \gets$ \choice{\dleft, \dright}\;
}

\If{\nispowerop}{
\text{$v\gets$ $v_{\textrm{right}} + v_{\textrm{left}} + 1$}\;
$d \gets$ [$0, 0, 0$]\;
}
\textbf{return} ($d$, $v$)\;}}
\textbf{end}
\end{algorithm*}


We apply the recursive Algorithm \ref{alg:dimPen} for dimensional analysis, which traverses the tree in the most straightforward way, like the evaluation itself, starting at the root node.
It becomes apparent that the joker unit is only introduced into the tree by constants.
As Table~\ref{tab:my_label} indicates, these jokers are propagated through the tree by most of the functions.
Unit violations occur when operands with non-matching or non-joker units are added or subtracted, as well as for functions which require dimensionless inputs.
When a violation occurs, the violation counter is increased by one (see lines 15, 28, 33), and the true output unit of the operation returned. For addition and subtraction, one of the operand units is chosen randomly. 
For example, the term $\log([1,2,0])$ violates the rules defined in Table \ref{tab:my_label}.
In this case, the true output unit [$0, 0, 0$] of the operation is returned.

When the traversal is complete, the algorithm returns the output unit $d$ of the equation as well as the number of unit violations $v$.
The Manhattan distance between $d$ and the target unit $d'$ is added to $v$ to also account for mismatches with the target unit.
A joker output is assumed to be equal to the target unit.

\subsection{Techniques to Handle Unit Violations in Symbolic Models}
Derived from the literature review in Sec.~\ref{sec:unitawareGP}, we introduce three techniques to deal with unit violations in GP trees.

\subsubsection{Evolutive culling}
The dimensional analysis is computationally cheaper compared to the fitting of constants followed by the numerical evaluation.
Evolutive culling makes use of this fact by performing the dimensional analysis directly after an offspring is created.
Individuals with unit violations, will be excluded from the population. Compared to the death penalty approach, this method saves time by avoiding fitting and evaluating an invalid model that will not survive the next generation because of the high penalty given to the primary objective.
Thus, the space of valid individuals can be explored more thoroughly.
As a  potential disadvantage, individuals with high accuracy but small unit violations cannot evolve into individuals without unit violations.
This might lead to overall worse performance regarding the primary objective.

\subsubsection{Repair mechanism}

For many fundamental laws of physics, constants alongside their units had to be discovered empirically to fit experimental observations. 
These multiplicative constants often have unconventional units, which balance output units of an equation to match the target unit. 
Vice versa, one can see a unit violation as a hint where such a balancing constant should be inserted.

We propose the following repair mechanism: whenever a unit violation occurs, a multiplicative constant is inserted into the tree at that  position to match the expected unit of the function.
Algorithm \ref{alg:dimPen} is modified so that a multiplicative constant is inserted whenever a unit violation is detected, rather than increasing the violation counter.
For example, an addition of [\unit{\metre}] and [\unit{\second}] can be balanced by multiplying one of the operands with a constant.
This turns the term into a joker so that the function returns the unit of the other operand according to Table~\ref{tab:my_label}.
The operand to be repaired is chosen randomly, so it can make a big difference how many constants are inserted depending on which operand is chosen.
When functions expecting dimensionless input receive an incorrect unit, the input term is multiplied by a constant to make it dimensionless.

Since the repair function is applied immediately after the offspring generation, only valid individuals are considered.
The repaired trees will then go into the fitting and evaluation process.
As a potential downside, the repair mechanism can lead to the insertion of many or unnecessary constants, which might negatively affect the primary objective and slows down the fitting process.

\subsubsection{Multi-objective approach}

The two methods discussed previously focus on exploring the space of valid, physics-adherent equations.
The multi-objective approach presented here allows for unit-violating individuals within the population, and it considers the number of unit violation as an additional objective.

Multi-objective optimization makes use of the concept of Pareto-dominance. 
Modern GP algorithms minimize multiple objectives at the same time, usually an error and a complexity objective.
Depending on the application, it can be beneficial to include a correlation measure as a supporting objective.
This helps individuals with poor accuracy but high correlation with the target variable to advance to the next generation, where they can continue evolving to better individuals.
Formulating constraints as additional optimization objectives is a common approach in GP \cite{haiderShapeconstrainedMultiobjectiveGenetic2023a,keijzerDimensionallyAwareGenetic1999a,zilleASSESSMENTMULTIOBJECTIVECOEVOLUTIONARY2021}.
We employ the NSGA-II algorithm to optimize multiple objectives simultaneously \cite{debFastElitistMultiobjective2002}.
The PO front contains multiple equations of the same level of complexity — with and without unit violations.
Equations without unit violations are preferred over equations with unit violations if they have the same accuracy and complexity.
However, there is no guarantee that a model without unit violations will be found for each level of complexity.

All algorithms are implemented in \texttt{TiSR}, a GP-based framework for thermo-dynamics-informed symbolic regression \cite{martinekIntroducingThermodynamicsInformedSymbolic2023a} written in Julia.
Its applicability is not limited to thermodynamics, but any kind of problems from the physics and engineering domain.
\texttt{TiSR} allows for fast algorithmic prototyping through simple code structures, while including all state-of-the-art components of a GP-based SR framework.

\section{Datasets and Experiment Configurations}

\subsection{Datasets}

The proposed algorithms are evaluated on known empirical equations from the \texttt{empiricalBench} benchmark presented in \cite{cranmerInterpretableMachineLearning2023}.
This benchmark does not include constants in the datasets so that the algorithms have to recover them alongside the form of the target equation.
Table \ref{tab:benchmarks} gives an overview of selected datasets for which dimensional analysis can be performed.
In addition, we use datasets from physics applications without ground truth.
The fluid mechanics dataset from the application of particle-laden flows was introduced by the authors in~\cite{reuterGraphNetworksInductive2023}.
A force~$F$ on a particle is computed from the positions of its neighboring particles in spherical coordinates~$r, \theta, \varphi$.
The thermodynamics dataset uses temperature~$T$ and density~$\rho$ of a gas mix to predict the pressure~$P$ \cite{VonPreetzmannKleinrahmEckmannCavuotoRichter:2021:1}.

\begin{table}[t]
    \centering
    \caption{Benchmark equations employed for our experiments with their input and target features and the respective units.}
    \begin{tabularx}{\textwidth}{>{\hsize=.52\hsize\linewidth=\hsize}X|>{\hsize=.4\hsize\linewidth=\hsize}X|>{\hsize=.6\hsize\linewidth=\hsize}X|>{\hsize=.42\hsize\linewidth=\hsize}X}
        Name & Equation & Input Features \& Units & Target Unit\\  \hline
        Hubble's Law & $v = H_0 D$ & Distance $D$ [\unit{\metre}] & Velocity $v$ [\unit{\metre\per\second}] \\ \hline
        Kepler's Third Law & $P = (\circ) \sqrt{a^3}$ & Distance $a$ [\unit{\metre}] &  Period $P$ [\unit{\day}] \\ \hline
        \multirow{2}{*}{Newton's Gravitation} & \multirow{2}{*}{$F = G \frac{m_1 m_2}{r^2}$} & Mass $m_1, m_2$ [\unit{\kilogram}], & \multirow{2}{*}{Force $F$ [\unit{\newton}]} \\
        & & Distance $r$ [\unit{\metre}] & \\ \hline
        \multirow{3}{*}{Ideal Gas Law} &  \multirow{3}{*}{$P = \frac{nRT}{V}$} & Number density $n$ [\unit{\mol}], &  \multirow{3}{*}{Pressure $P$ [\unit{\pascal}]} \\
        & & Temperature $T$ [\unit{\kelvin}], & \\
        & & Volume $V$ [\unit{\metre\cubed}] & \\ \hline
        Rydberg Formula &  $\lambda = \frac{1}{R_H(\frac{1}{n_1^2} - \frac{1}{n_2^2})}$ & Principal Quantum Number $n_1, n_2$ [$\cdot$] & Wavelength $\lambda$ [\unit{\metre}] \\ \hline
        \multirow{2}{*}{Fluid Mechanics} & \multirow{2}{*}{unknown} & Distance $r$ [\unit{\metre}], & \multirow{2}{*}{Force $F$ [\unit{\newton}]}\\
         & & Angle $\theta, \varphi$ [$\cdot$] & \\ \hline
         \multirow{2}{*}{Thermodynamics} &  \multirow{2}{*}{unknown} & Temperature $T$ [\unit{\kelvin}], & \multirow{2}{*}{Pressure $P$ [\unit{\pascal}]}\\
         & & Density $\rho$ [\unit{\kilogram\per\metre\cubed}] & \\
    \end{tabularx}
    \label{tab:benchmarks}
\end{table}

We also study the sensitivity of the proposed algorithms to noise.
When recovering the exact equation on noisy data, the choice of the noise level is an important parameter.
It has to be guaranteed that the noisy data is still described best by the target equation, and not a different one of the same complexity.
We assume that beyond $10\%$ noise, it is difficult to recover the exact equation.
The noise levels of $5\%$ and $10\%$ were inspired by~\cite{defrancaInterpretableSymbolicRegression2023,lacavaContemporarySymbolicRegression2021}.
For the Rydberg equation, noise levels beyond $3\%$ were too noisy for the exact equation to be recovered, as experiments with $10\%$ noise indicated \cite{cranmerInterpretableMachineLearning2023}.
We thus applied $1\%$ and $3\%$ noise.

\subsection{Experiment Configurations}

Table~\ref{tab:algoConfig} gives an overview of the algorithm settings and use-case dependent function sets.
The input features and units from Table~\ref{tab:benchmarks} are the training data of the algorithms, so that necessary constants need to be identified by the algorithms.
For other parameters, the standard settings of \texttt{TiSR} are used \cite{martinekIntroducingThermodynamicsInformedSymbolic2023a}.
We set time limits of thirty minutes for experiments on \texttt{empiricalBench} datasets and sixty minutes for experiments without ground truth. 
This approach is favored over fixed generation counts due to algorithmic modifications that affect the generation runtime. 
However, we aim to evolve unit-adherent equations without compromising runtime efficiency.
We compare the proposed algorithm to a baseline algorithm without dimensional analysis.
All algorithms optimize multiple objectives at the same time: the mean squared error (MSE), the function complexity and the Spearman correlation as a supporting objective as defined in \cite{zilleASSESSMENTMULTIOBJECTIVECOEVOLUTIONARY2021}.
In addition, we assess an algorithm that minimizes the number of unit violations as a fourth objective.
Each algorithm is repeated 31 times.

\begin{table}[t]
    \centering
    \caption{Algorithm Configurations for Experiments}
    \begin{tabular}{l|l}
        Population size & 500 \\
        Max. complexity of equations & 30 \\
        Complexity of variables and functions & 1 \\
        Complexity of constants & 2 \\
        Function set empiricalBench & {$+, -, \cdot, \div, e^{\circ}, \log(\circ), \sqrt{\circ}, \circ^2, \circ^3$} \\
        Function set Fluid Mechanics & {$+, -, \cdot, \div, e^{\circ}, \log(\circ), \sin(\circ), \cos(\circ), \circ^{\circ}$}\\
        Function set Thermodynamics & {$+, -, \cdot, \div, e^{\circ}, \log(\circ), \circ^{\circ}$}
    \end{tabular}
    \label{tab:algoConfig}
\end{table}

\subsection{Evaluation Procedure}

The assessment whether an algorithm identified a specific target equation correctly comes with two major issues: first, the selection of a solution from the PO front.
Finding the best trade-off between accuracy and complexity automatically is a complex task.
And second, the equivalence check of two equations using libraries like Python \texttt{sympy} or Julia \texttt{SymbolicUtils}.
As related studies report \cite{lacavaContemporarySymbolicRegression2021}, small differences in the simplification as well as the value of fitted constants might lead to misclassification.
To overcome these issues and base our analysis on trustworthy results, we eye-check each PO front for the target equation,  which makes a total of more than 1800 checked PO fronts.
Some parts of the analysis can be accelerated by automatically scanning a PO front for solutions which have already been classified as correct by a human.
An equation counts as solved when the shape of the equation is correct, the exact values of the fitted constants are irrelevant.
We define two stages of success: finding the exact solution and finding a solution close to the exact one, which is measured by eyeball.
For the datasets without ground truth, we analyze the PO fronts.

\section{Results and Analysis}

\subsection{Empirical Datasets with Known Solutions}
Table~\ref{tab:results_known} gives an overview of the performance of the proposed algorithms on known benchmark datasets of equations.
It becomes apparent that all algorithms recover the correct equations for all datasets and all noise levels with a high success rate.
Only the proposed repair mechanism has lower rates of identifying the exact equation as the noise level increases.
It still finds solutions close to the target equation in the final PO front, which often contain additional constants.
Overall, we conclude from these results that evolutive culling as well as the multi-objective approach perform at least as good as the baseline method.
However, it should be noted that there is almost no space for improvement, as the baseline algorithm finds the correct solution in almost all cases.

\begin{table}[t]
    \centering
    \caption{Number of correct/almost correct/wrong rediscoveries of target equations for different datasets and noise levels out of 31 runs.}
    \begin{tabular}{p{0.15\textwidth}<{\centering}|p{0.15\textwidth}<{\centering}|p{0.15\textwidth}<{\centering}|p{0.15\textwidth}<{\centering}|p{0.15\textwidth}<{\centering}|p{0.15\textwidth}<{\centering}}
        Dataset                   & Noise Level & Baseline & Evolutive Culling & Repair Mechanism & Multi-objective \\ \hline
        \multirow{3}{*}{Hubble}   & $0\%$       & 31/0/0   & 31/0/0            & 30/1/0           & 31/0/0 \\
                                  & $5\%$       & 31/0/0   & 31/0/0            & 28/3/0           & 31/0/0 \\
                                  & $10\%$      & 31/0/0   & 31/0/0            & 26/5/0           & 31/0/0 \\ \hline
        \multirow{3}{*}{Kepler}   & $0\%$       & 31/0/0   & 31/0/0            & 31/0/0           & 31/0/0 \\
                                  & $5\%$       & 31/0/0   & 31/0/0            & 25/6/0           & 31/0/0 \\
                                  & $10\%$      & 31/0/0   & 30/1/0            & 26/5/0           & 31/0/0 \\\hline
        \multirow{3}{*}{Newton}   & $0\%$       & 31/0/0   & 31/0/0            & 31/0/0           & 31/0/0 \\
                                  & $5\%$       & 31/0/0   & 31/0/0            & 31/0/0           & 31/0/0 \\
                                  & $10\%$      & 31/0/0   & 31/0/0            & 31/0/0           & 31/0/0 \\\hline
        \multirow{3}{*}{Ideal Gas}& $0\%$       & 31/0/0   & 31/0/0            & 31/0/0           & 31/0/0 \\
                                  & $5\%$       & 30/1/0   & 31/0/0            & 16/17/0          & 31/0/0 \\
                                  & $10\%$      & 31/0/0   & 31/0/0            & 8/23/0           & 31/0/0 \\\hline
        \multirow{3}{*}{Rydberg}  & $0\%$       & 31/0/0   & 31/0/0            & 31/0/0           & 29/0/2 \\
                                  & $1\%$       & 31/0/0   & 31/0/0            & 31/0/0           & 31/0/0 \\
                                  & $3\%$       & 27/3/1   & 29/1/1            & 24/4/3           & 29/1/1 \\
    \end{tabular}
    \label{tab:results_known}
\end{table}

\subsection{Empirical Datasets with Unknown Solutions}

For the thermodynamics (TD) and fluid mechanics (FM) datasets, no ground truth solution is known.
To compare the algorithms, we analyze the resulting PO fronts for interesting characteristics: the numbers of solutions, the percentage of solutions with unit violations, and the mean number of constants in the equation.
Furthermore, we look at the number of generations performed within the time limit.
For pairwise statistical comparison to the baseline method, the non-parametric Mann-Whitney U test at a confidence level $\alpha = 0.95$ is performed.

\begin{figure}[t]
    \centering
    \caption{Measurements on the Pareto-optimal front for datasets with unknown solutions from thermodynamics and fluid mechanics over 31 independent runs.}
    \includegraphics[width= 0.99\textwidth]{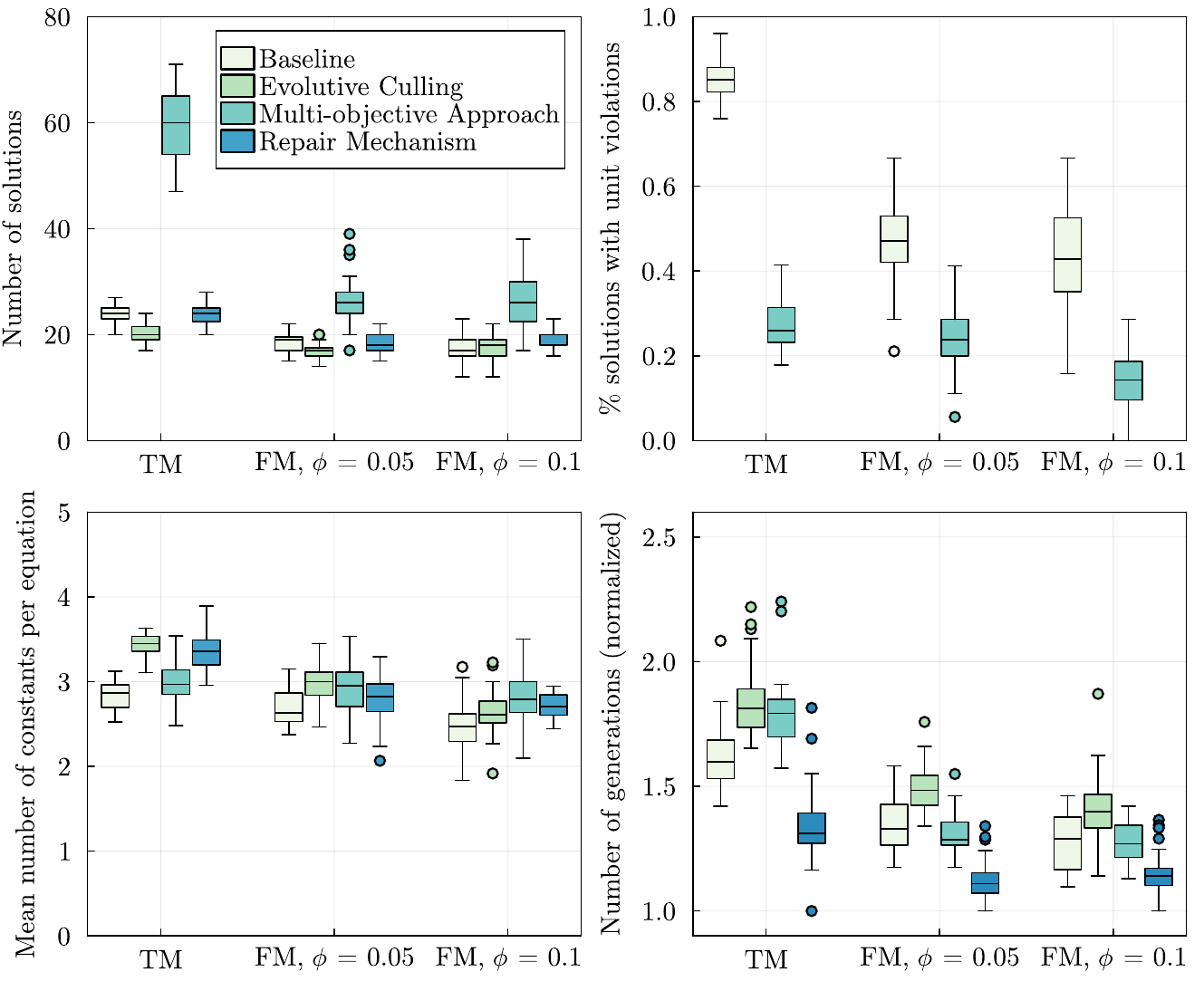}
    \label{fig:results_unknown}
\end{figure}

Figure~\ref{fig:results_unknown} indicates that the PO fronts of the multi-objective approach contain more solutions compared to the other approaches, which is supported by statistical tests.
This can be explained with the additional unit violation objective, which allows the algorithm to include solutions with multiple levels of unit violations per complexity value.

The upper right subplot of Figure~\ref{fig:results_unknown} shows the percentage of solutions with unit violations within the PO front.
When using the multi-objective algorithm, if a complexity level has multiple solutions with varying numbers of unit violations, the lowest one is selected.
If this is zero, no unit violations are considered for that complexity level.
First, we observe that it can indeed be problematic to exclude dimensional analysis from the algorithm when the requirement for unit-adherent equations exists.
This is reflected by median values of more than $80\%$ of solutions with unit violations on the TD dataset and more than $40\%$ on the FM datasets when the baseline algorithm without dimensional analysis is applied.
The multi-objective approach not only finds more solutions, but also more solutions without unit violations.
The difference is particularly drastic for the TD dataset, but can also be observed for the FM datasets.
Evolutive culling and the repair mechanism ensure unit-compliance of the equations, resulting in a final front with $0\%$ of solutions with dimensional error.
On this criterion, all proposed methods outperform the baseline algorithm with statistical significance.

The number of constants within an equation is an important quality criterion for domain experts, who prefer models with fewer constants.
On the TD dataset, evolutive culling and repair mechanism contain equations with significantly more constants in the PO front than the baseline algorithm.
This cannot be confirmed statistically for the FM datasets, but a similar tendency can be observed in the bottom left subplot of Figure~\ref{fig:results_unknown}.
Evolutive culling does not insert new constants into the tree like the repair mechanism does, but still shows higher usage of constants.
This can be explained by the joker unit, which is introduced only by constants and propagated through the tree by most functions, encouraging the use of constants in equations.
The multi-objective approach does not show significant differences to the baseline in the numbers of constants.

By looking at the number of generations completed within the time limit, we aim to assess the runtime differences between the algorithms.
The number of generations is normalized by the minimum number of generations a single run achieved within each dataset to account for different dataset sizes.
It can be seen that evolutive culling tends to run more generations and the repair mechanism runs fewer generations compared to the baseline algorithm.
These observations are supported by the results of the statistical test.
The runtime loss of the repair mechanism can be explained with the higher number of constants that need to be fitted, which increases the duration of one generation.
Evolutive culling excludes solutions with unit violations from the population, which leads to smaller population sizes in the current implementation of the algorithm.
This explains the higher number of generations performed by the algorithm.
The multi-objective approach performs significantly more generations on the TD dataset, which is not continued for the FM datasets.
A more profound understanding of this behavior requires a closer look at the population dynamics during the evolution.
\begin{figure}[t]
    \centering
    \caption{Solutions of 31 combined PO fronts per algorithm on the TD dataset. The magnitude of unit violations is color-coded from white (0 violations) to black (22 violations), with 22 being the maximum number of unit violations on the TD dataset.}
    \includegraphics[width= 0.99\textwidth]{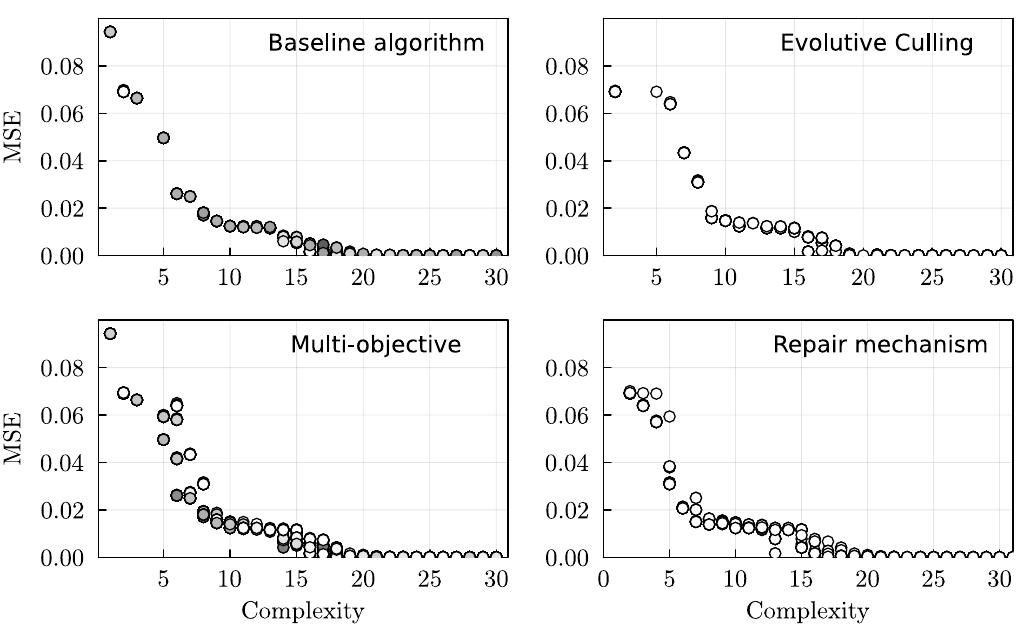}
    \label{fig:results_POfronts}
\end{figure}

Figure~\ref{fig:results_POfronts} displays the 31 combined PO fronts for each algorithm on the TD dataset.
We seek to examine the effects of the algorithms with dimensional analysis on the primary error objective MSE.
Unsurprisingly, the baseline algorithm contains more solutions with unit violations in the PO front.
Looking at the multi-objective approach, one can see that the unit-adherent solutions with complexities between five and eight have considerably higher MSE values than the ones with unit violations.
This effect almost vanishes for higher complexities from nine to 15.
For complexities above 16, all algorithms identify solutions with MSE values close to 0.
The algorithms with dimensional analysis thereby have fewer unit violations than the baseline.
Due to space reasons, we only analyze the TD dataset here.
Similar observations are made for the FM datasets.

\section{Conclusions and Future Work}

We applied a method for unit-aware GP that includes constants with undetermined units.
Constants introduce “joker” units, which are propagated through the tree according to a propagation scheme.
The dimensional analysis returns the magnitude of unit violations of an equation.
Two approaches were presented to avoid unit-violating individuals during evolution: evolutive culling and a repair mechanism.
The additionally proposed multi-objective approach minimizes the magnitude of unit violations as an additional objective.
Experiments conducted on datasets of known equations have shown that both evolutive culling and the multi-objective approach perform as well as a baseline method without dimensional analysis.
The repair mechanism often introduces more constants than necessary, which is an undesired behavior.
No advantages of unit-aware approaches were observed compared to the baseline method when the noise level in the datasets increased.
This indicates that more complex benchmark equations should be employed in the future.
In-depth analysis of the PO fronts for benchmark datasets without ground truth revealed that a large share of solutions in the PO front of the baseline algorithm have unit violations.
All proposed unit-aware algorithms were able to identify solutions with similarly low error but without unit violations. However, evolutive culling and the repair mechanism showed higher usage of constants compared to the baseline algorithm.

When the requirement for unit-adhering equations exists, it is definitely beneficial to include unit information in the GP algorithm.
The proposed algorithms have shown low sacrifices in accuracy on the used datasets.
From a practical perspective, we prefer the multi-objective approach as it offers decision makers multiple levels of unit violations per complexity.
However, to better understand the strengths and weaknesses of each algorithm, further investigation on the population dynamics using more complex benchmark equations is necessary. 
One could also think of combining the methods, such as repairing solutions with small dimension error, culling solutions with large dimension error, and using the multi-objective unit-aware approach as an overall optimization algorithm.
\subsubsection{\ackname} This work was funded by the Deutsche Forschungsgemeinschaft (DFG, German Research Foundation) —  MO~1792/11-1 and HE~6077/14-1 — within the Priority Programme “SPP 2331: Machine Learning in Chemical Engineering”.

%
%

\newpage
\bibliographystyle{splncs04}
\bibliography{PPSN2024}
\end{document}